\documentclass[letterpaper, 10 pt, conference]{ieeeconf}  % Comment this line out
\IEEEoverridecommandlockouts                              % This command is only
\title{Approximating Discontinuous Nash Equilibrial Values of Two-Player General-Sum Differential Games}

\author{Lei Zhang$^{1}$, Mukesh Ghimire$^{1}$, Wenlong Zhang$^{2}$, Zhe Xu$^{1}$, Yi Ren$^{1}$% <-this % stops a space
\thanks{This work was supported by the National Science Foundation under Grant CMMI-1925403.}
\thanks{$^{1}$L. Zhang, M. Ghimire, Z. Xu, and Y. Ren are with Department of Mechanical and Aerospace Engineering, Arizona State University, Tempe, AZ, 85287, USA. Email:
        {\tt\small \{lzhan300, mghimire, xzhe1, yiren\}@asu.edu}}%;
\thanks{$^{3}$W. Zhang is with The Polytechnic School, Ira A. Fulton Schools of Engineering, Arizona State University, Mesa, AZ, 85212, USA. Email:
        {\tt\small wenlong.zhang@asu.edu}}%   
}

\usepackage[utf8]{inputenc}
\usepackage{amssymb}
\usepackage{amsmath}
\usepackage{graphicx}  %Required
\usepackage{xcolor}
\usepackage[colorlinks=true]{hyperref}
\usepackage[noadjust]{cite}
\DeclareMathOperator*{\argmax}{arg\,max}

\usepackage[linesnumbered,ruled]{algorithm2e} 
\usepackage{color}

\usepackage{graphicx}       
\usepackage{wrapfig}
\usepackage{subfigure}     

\usepackage{booktabs}
\usepackage{dsfont}
\usepackage{bbm}
\usepackage{caption}
\usepackage{multirow}
\usepackage[font=footnotesize]{caption}

\iftrue % use space-saving macro
        \newcommand{\cutsectionup}{\vspace*{-0.1in}}

        \newcommand{\cutparagraphup}{\vspace*{-0.17in}}
        \newcommand{\cutparagraphdown}{\vspace*{-0.03in}}

        \newcommand{\cutcaptionup}{\vspace*{-0.1in}}
        \newcommand{\cutcaptiondown}{\vspace*{-0.2in}}

        \newcommand{\cutequationup}{\vspace*{-0.07in}}
        \newcommand{\cutequationdown}{\vspace*{-0.07in}}

        \newcommand{\cuttableup}{}
        \newcommand{\cuttabledown}{}

        \newcommand{\cut}{{\vspace*{-0.02in}}}
        \newcommand{\cutmore}{{\vspace*{-0.06in}}}
        \newcommand{\negcut}{}

\else % do not use space-saving macro
        \newcommand{\cutsectionup}{}

        \newcommand{\cutparagraphup}{
        \newcommand{\cutparagraphdown}{}

        \newcommand{\cutcaptionup}{}
        \newcommand{\cutcaptiondown}{}

        \newcommand{\cutequationup}{}
        \newcommand{\cutequationdown}{}

        \newcommand{\cuttableup}{}
        \newcommand{\cuttabledown}{}

        \newcommand{\cut}{}
        \newcommand{\cutmore}{}
        \newcommand{\negcut}{}
\fi

\begin{document}
\maketitle
\thispagestyle{empty}
\pagestyle{empty}

\begin{abstract}
Finding Nash equilibrial policies for two-player differential games requires solving Hamilton-Jacobi-Isaacs (HJI) PDEs. Self-supervised learning has been used to approximate solutions of such PDEs while circumventing the curse of dimensionality. However, this method fails to learn discontinuous PDE solutions due to its sampling nature, leading to poor safety performance of the resulting controllers in robotics applications when player rewards are discontinuous. This paper investigates two potential solutions to this problem: a hybrid method that leverages both supervised Nash equilibria and the HJI PDE, and a value-hardening method where a sequence of HJIs are solved with a gradually hardening reward. We compare these solutions using the resulting generalization and safety performance in two vehicle interaction simulation studies with 5D and 9D state spaces, respectively. 
% extends from previous SOTA on zero-sum games with continuous values to general-sum games with discontinuous values, where the discontinuity is caused by that of the players' losses. We show that due to its lack of convergence proof and generalization analysis on discontinuous losses, the existing self-supervised learning technique fails to generalize and raises safety concerns in an autonomous driving application.
Results show that with informative supervision (e.g., collision and near-collision demonstrations) and the low cost of self-supervised learning, the hybrid method achieves better safety performance than the supervised, self-supervised, and value hardening approaches on equal computational budget. Value hardening fails to generalize in the higher-dimensional case without informative supervision. Lastly, we show that the neural activation function needs to be continuously differentiable for learning PDEs and its choice can be case dependent.
% : Among \texttt{relu}, \texttt{sin}, and \texttt{tanh}, we show that \texttt{tanh} is the only choice that achieves optimal generalization and safety performance. Our conjecture is that \texttt{tanh} (similar to \texttt{sin}) allows continuity of value and its gradient, which is sufficient for the convergence of learning, and at the same time is expressive enough (similar to \texttt{relu}) at approximating discontinuous value landscapes. 
% Lastly, we apply our method to approximating control policies for an incomplete-information interaction and demonstrate its contribution to safe interactions.
\end{abstract}

% \cutsectionup
\section{Introduction}
% \cutsectiondown
\label{sec:intro}

\textbf{Problem statement.} Human-robot interactions (HRIs) in real time can be modeled as general-sum differential games.
% which are different from zero-sum games and model situations in that players achieve competitive-not necessarily opposite-objectives~\cite{karlin2017game,fridovich2020efficient}. 
The Nash equilibrial value of the game is a viscosity solution to the Hamilton-Jacobi-Isaacs (HJI) equations~\cite{viscosity}, and is a function of the players' states and time. 
% Within the presented context, this paper focuses on the task of approximating the Nash equilibrial value function of a two-player general-sum differential game with complete information. 
Conventional algorithms for solving Hamilton-Jacobi PDEs are known to suffer from the curse of dimensionality~\cite{mitchell2003}, i.e., values become hard to compute for high-dimensional state spaces. Recent attempts consider self-supervised (i.e., physics-informed) learning that forces the consistency between the value and the PDE, and have achieved empirical success on a variety of differential games~\cite{deepreach,shin2020convergence}. While convergence of this method has been proven for Lipschitz and H\"{o}lder continuous player rewards~\cite{mitchell2005,shin2020convergence}, our experiments suggest that convergence to the true values cannot be achieved when players' rewards, and thus the values, are discontinuous with respect to states and time. 
Such discontinuity can occur when players receive both continuous rewards, e.g., energy consumption or path-following losses, and those encoded from temporal logic, e.g., safety penalties. We discuss the challenge of learning discontinuous values using a toy case in Sec.~\ref{sec:methods}.

\textbf{Solutions.} We examine two potential solutions to this challenge: %(1) A supervised method where value is learned based on sample equilibrial trajectories derived from Pontryagin's Maximum Principle (PMP)~\cite{mangasarian1966sufficient}, (2) a hybrid method where both supervised data and the HJI equations are used, 
(1) A hybrid method where we leverage both supervised equilibrial data generated by Pontryagin's Maximum Principle (PMP)~\cite{mangasarian1966sufficient} and the HJI equations; (2) A modified self-supervised method where we gradually harden an initially softened reward. Comparisons are performed on two vehicle interaction studies: An uncontrolled intersection case with a 5D value function with complete information where players know each other's types and incomplete information where player types are private and inferred, and a collision avoidance case with a 9D value function. 

\textbf{Contributions.} Our study leads to the following new findings: 
(1) We show that the hybrid method achieves better generalization performance on value prediction and safer control policies than other methods under the same computational budget. On the other hand, value-hardening fails to generalize in the higher-dimensional case, indicating the importance of informative supervision. To the authors' best knowledge, this is the first paper that addresses the challenge in approximating values of HJI PDEs with discontinuous reward; (2) We show that the choice of neural activation in the value network is important: \texttt{relu} does not generalize well due to its discontinuous derivative. Performance of \texttt{sin} activation varies across case studies, indicating the need for hyperparameter tuning. \texttt{tanh} and continuously differentiable variants of \texttt{relu}, e.g., \texttt{gelu}, generalize well using the hybrid method. These new findings add counter examples to the existing literature that promotes the use of \texttt{sin} activation for value approximation~\cite{deepreach}, and support recent discussions on the need for adaptive activations for neural network-based PDE solvers~\cite{jagtap2020adaptive}.

\section{Related Work}
\label{sec:related}

% \citet{wiggers2016structure} gives insight on how the value function of a zero-sum partially observable stochastic game generalizes over the space of sufficient statistics through POMDP in incomplete-information settings. 
\textbf{Solving Hamilton-Jacobi PDEs via deep learning.} 
With the development of auto-differentiation~\cite{autodiff}, deep neural networks have become means to solving PDEs when analytical methods suffer from the curse of dimensionality~\cite{bellman1965dynamic}. Existing methods of this kind can be grouped into three types: ``Boundary matching'' methods first reformulate PDEs as backward stochastic differential equations and solve them by forcing a match between the terminal states of the resultant stochastic processes and the boundary conditions~\cite{bsde, han2020convergence}. ``Equation matching'' methods, which are often called physics-informed neural nets (PINN), directly force a match between a surrogate function and the PDE~\cite{ito2021neural,deepreach}. Both boundary and equation matching are self-supervised, and have proven convergence to the ground truth solutions when both the network and the governing equations are continuous~\cite{han2020convergence,ito2021neural,shin2020convergence}. Lastly, supervised learning methods train a network based on sampled solutions to the PDE. To the authors' best knowledge, there is currently few generalization analysis for this method. Recent studies have investigated the effectiveness of ``equation matching'' at solving PDEs with discontinuous solutions (e.g., Burgers equation~\cite{jagtap2020adaptive}) where both initial and terminal boundaries are specified. We show in this paper that PDEs with only terminal (or initial) boundary conditions, such as HJIs, cause an unidentifiability issue. 

% Bansal et al.~\citep{deepreach} used sinusoidal representation networks (SIREN)~\citep{siren} to approximate values governed by HJI Variational Inequalities (HJI VIs) using a self-supervised approach, and demonstrated its favorable scalability against the level-set solver~\citep{mitchell2005} at approximating  backward reachable sets. 
% The convergence of this algorithm is analyzed in \citep{shin2020convergence, ito2021neural} under the assumption of continuous losses. 

\textbf{Games with incomplete information.} 
Zero-sum dynamic games with incomplete- or imperfect-information can be solved by counterfactual regret minimization (CFR)~\cite{zinkevich2007regret} or Regularized Nash Dynamics (RND)~\cite{perolat2022mastering}, which have been successfully applied to heads-up Texas hold'em poker~\cite{tammelin2015solving,facebook_poker} and Stratego~\cite{perolat2022mastering}.
% and more recently to the no-limit version~\citep{facebook_poker} when combined with subgame solving~\citep{subgame} and search methods~\citep{brown2018depth}. 
However, there have so far been few applications of CFR and RND to continuous-time interactions with incomplete information, e.g., through time discretization. 
%due to the lack of scalability of the algorithm with respect to state and belief dimensions and time~\citep{***}. 
On the other hand, attempts to directly solve HRIs are often facilitated by simplifications of the underlying games that balance theoretical soundness and practical utility. 
% Card and video games can be as POMDP games and solved by CFR, Markov Chain Monte Carlo (MCMC) and DRL~\citep{xiang2021recent}.
Most attempts of this type simplify games as optimal control problems or complete-information games (e.g., \cite{foerster_learning_2017,sadigh_planning_2018,kwon2020humans,schwarting2019social,zahedi2022seeking,li2022off}), and some use belief updates to adapt motion planning (e.g., \cite{nikolaidis2017human,sun2018probabilistic,peng2019bayesian,wang2019enabling,fridovich2020confidence,chen2021shall}). 
For real-time control, \cite{fridovich2020efficient} proposed to solve linear-quadratic games iteratively to approximate the equilibrial policy of a complete-information differential game. Unlike existing studies, in this paper we achieve real-time belief update and feedback control in incomplete-information games by learning off-line the Nash equilibrial values of hypothetical games to be played by all possible combinations of player types. The values then enable online Bayesian belief update to correct players' (false) beliefs about each other's type.
% It is yet an open question as in what conditions simplified models of games lead to policies that approximate perfect Bayesian equilibrium of incomplete-information games.  

\cutsectionup
\section{Methods}
\label{sec:methods}
\textbf{Notations.} Let $\mathcal{X}_i$ (resp. $\mathcal{U}_i$) be the state (resp. action) space for Player $i$.
% , and $\Theta$ the shared set of player types. 
The time-independent state dynamics of Player $i$ is denoted by $\dot{x}_i[t] = f(x_i[t], u_i[t])$ for $x_i[t] \in \mathcal{X}_i$ and $u_i[t] \in \mathcal{U}_i$. Dependence of time will be omitted when possible. For a fixed-time interaction within $[0, T]$, the instantaneous and terminal losses of Player $i$ are denoted by $l((x_i, x_{-i}), u_i)$ and $c(x_i, x_{-i})$, respectively. 
% We assume shared functions $f(\cdot, \cdot)$, $l(\cdot, \cdot; \cdot)$, and $c(\cdot, \cdot; \cdot)$ for all players. 
For a complete-information differential game between two players, the value function of Player $i$ is $\nu_i(\cdot, \cdot, \cdot): \mathcal{X}_i \times \mathcal{X}_{-i} \times [0,T] \rightarrow \mathbb{R}$ where the arguments are in the order of the ego's (Player $i$'s) states, the other player's states, and the current time. To reduce notational burden, we will use $f_i$, $l_i$, $c_i$, and $\nu_i$ respectively for the player-wise dynamics, losses, and the value, and use $\textbf{a}_i=(a_i, a_{-i})$ to concatenate variables from the ego ($i$) and the other players. We use $\nabla_x\cdot$ to denote partial derivative with respective to $x$.

\textbf{Preliminaries.} \textit{Hamilton-Jacobi-Isaacs equations}: The Nash equilibrial values of a two-player general-sum differential game solve the following HJI equations ($L$) and satisfy the boundary conditions ($D$)~\cite{starr1969nonzero}:
\begin{equation}
\begin{aligned}
& L(\nu_i, \nabla_{\textbf{x}_i} \nu_i, \textbf{x}_i, t) := \nabla_t \nu_i + \nabla_{\textbf{x}_i} \nu_i ^T \textbf{f}_i - l_i = 0 \\
& D(\nu_i, \textbf{x}_i) := \nu_i(\textbf{x}_i, T) - c_i = 0, \quad  \text{for} \quad i = 1, 2. \\
\end{aligned}
\label{eq:hji}
\end{equation}
Here players' policies are derived by maximizing the equilibrial Hamiltonian $h_i(\textbf{x}_i, \nabla_{x_i} \nu_i, t) = \nabla_{x_i} \nu_i^T f_i - l_i$: $u_i = \argmax_{u \in \mathcal{U}_i} \left\{h_i\right\}$ for $i = 1, 2$.
% \begin{equation}
% u_i = \argmax_{u \in \mathcal{U}_i} \left\{h_i\right\} \quad  \text{for} \quad i = 1, 2.
% \label{eq:hamiltonian}
% \end{equation}

\textit{Pontryagin's Maximum Principle}: While solving the HJI would provide a feedback control policy, it is often more tractable to compute open-loop policies for specific initial state $(\bar{x}_1, \bar{x}_2) \in \mathcal{X}_1 \times \mathcal{X}_2$ by solving the following boundary value problem (BVP) according to PMP\footnote{We note that solving BVP has its own numerical challenges when the equilibrium involves singular arcs~\cite{singarcs}. These challenges are not explored in this paper.}:
\begin{equation}
\begin{aligned}
& \dot{x}_i = f_i, ~\quad x_i[0] = \bar{x}_i, \\
& \dot{\lambda}_i = - \nabla_{x_i} h_i, ~\quad \lambda_i[T] = - \nabla_{x_i} c_i, \\
& u_i = \argmax_{u \in \mathcal{U}_i} ~\{h_i\}  \quad \text{for} \quad i = 1, 2.
\end{aligned}
\label{eq:pmp}
\end{equation}
Here $\lambda_i$ is the time-dependent co-state for Player $i$. The co-state connects PMP and HJI through $\lambda_i = \nabla_{x_i} \nu_i$. Solutions to Eq.~\eqref{eq:pmp} are specific to the given initial states.

\textbf{Self-supervised learning for differential games.} This approach directly trains a neural network that approximately satisfies the governing equations. 
% This approach is also called physics-informed machine learning~\cite{shin2020convergence} for the reason that it directly relies on the governing equations rather than sampled solutions of such equations.
Let $\hat{\nu}_i(\cdot,\cdot,\cdot): \mathcal{X}_i \times \mathcal{X}_{-i} \times [0,T] \rightarrow \mathbb{R}$ be a neural network that approximates $\nu_i$, and let $\mathcal{D} = \left \{\left(x_1^{(k)}, x_2^{(k)}, t^{(k)} \right) \right\}_{k=1}^K$ be uniform samples in $\mathcal{X}_1 \times \mathcal{X}_1 \times [0, T]$.
We extend the existing formulation for solving zero-sum games~\cite{deepreach} to the general-sum setting:
% \begin{equation}
% \min_{\hat{\nu}_1, \hat{\nu}_2} \quad L_2\left(\hat{\nu}_1, \hat{\nu}_2; \theta \right) := \sum_{k=1}^K \sum_{i=1}^2 \left\|L(\hat{\nu}_i, \nabla_{\textbf{x}_i} \hat{\nu}_i, \textbf{x}_i^{(k)}, t^{(k)})\right\| + C_2 \phi\left(D(\hat{\nu}_i, \textbf{x}_i^{(k)})\right),
% \label{eq:deepreach}
% \end{equation}
\cutequationup
\begin{equation}
\begin{aligned}
    \min_{\hat{\nu}_1, \hat{\nu}_2} \quad L_1\left(\hat{\nu}_1, \hat{\nu}_2; \theta \right) := & \sum_{k=1}^K \sum_{i=1}^2 \left\|L(\hat{\nu}_i, \nabla_{\textbf{x}_i} \hat{\nu}_i, \textbf{x}_i^{(k)}, t^{(k)})\right\|\\
    & + C_1 \phi\left(D(\hat{\nu}_i, \textbf{x}_i^{(k)})\right),
    \label{eq:deepreach}
\end{aligned}
\end{equation}
where $\hat{\nu}_i^{(k)}$ is a shorthand for $\hat{\nu}_i \left(x_i^{(k)}, x_{-i}^{(k)}, t^{(k)} \right)$ and $C_1$ balances the inconsistencies of the value network to the HJI ($L$) and to the boundary conditions ($D$). It is worth noting that in each iteration of solving Eq.~\eqref{eq:deepreach}, a sub-routine is needed to find the control policies that maximize the Hamiltonian. \cite{ito2021neural} proved the convergence of $\hat{\nu}_i$ to $\nu_i$ via solving Eq.~\eqref{eq:deepreach} when $\phi(\cdot)$ is a $H^{3/2}$-norm; \cite{shin2020convergence} provided convergence and generalization analyses of the learning in forms of Eq.~\eqref{eq:deepreach} for linear second-order elliptic and parabolic type PDEs (which include HJI) with continuity assumptions on $L$, $D$, and $\hat{\nu}_i$. 
% \cite{deepreach} empirically showed convergence of the learning on differential games when $\phi(\cdot)$ is an $L^1$-norm. We test $L^1$- and $L^2$-norms for $\phi(\cdot)$ in Sec.~\ref{sec:uncontrolled intersection}.

\begin{figure}
\centering
% \vspace{0.02in}
\includegraphics[width = 0.48\textwidth]{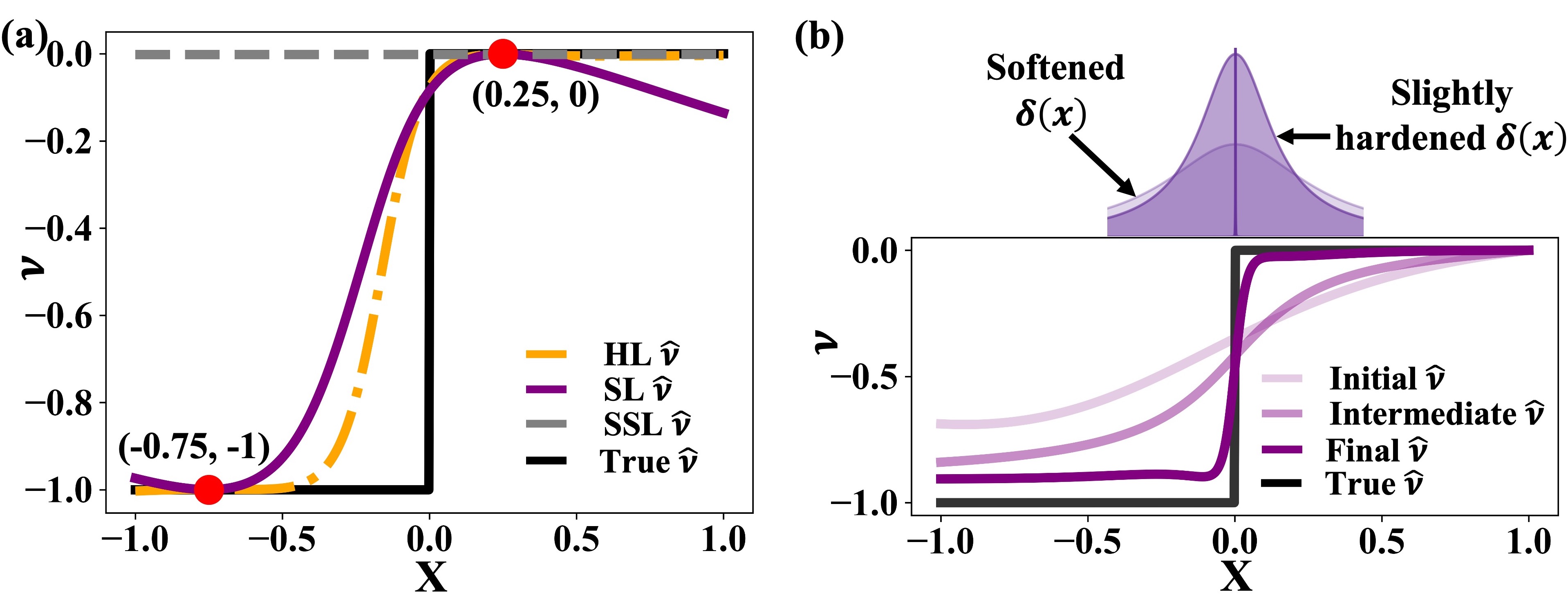}
\caption{(a) Value comparison among the learning methods for a simple 1D case. Red dots are the supervised data. (b) Evolution of the value function due to gradually hardening delta function. Delta functions are shown on top. Transparency reduces with hardening.}
\label{fig:toy_case}
\vspace{-20pt}
\end{figure}

% \begin{wrapfigure}{r}{0.2\textwidth}
% \centering
% \vspace{-10pt}
% \includegraphics[width = 0.2\textwidth]{toy_case.png}
% \caption{Value comparison among the learning methods for a simple 1D case. Red dots are the supervised data.}
% \vspace{-10pt}
% \label{fig:toy_case}
% \end{wrapfigure}

\textbf{Challenge in approximating discontinuous HJI values.} 
We use a toy case to explain the challenge in approximating discontinuous solutions of a differential equation with only terminal boundary conditions using self-supervised learning: Consider a 1D function $\nu(\cdot)$ as a solution to $\nabla_x \nu - \delta(x)=0$ with the boundary condition $\nu(1) = 0$ within $x \in [-1, 1]$, where $\delta(x)$ is a delta function that peaks at $x=0$ (see Fig.~\ref{fig:toy_case}).
With uniform samples, the self-supervised loss ($L_1$) can be minimized almost surely using a horizontal line $\hat{\nu}(x) = 0$. This is because due to the differential nature of the governing equation, the accuracy of a self-supervised model at one point in space and time solely relies on that of its neighboring samples, yet informative neighbors (here at $x = 0$) have low (here zero) probability to be sampled. We also note that when both initial and terminal boundaries are specified (here $\nu(-1)=-1$ and $\nu(1)=0$), self-supervised learning can work, e.g., in the case of a Burgers equation. 

\textbf{Solutions.} (1) \textit{Hybrid method}: Supervised data reveals the structure of the solution. In the toy case, an approximation to the solution can be learned based on two informative data points sampled from each side of $0$ (see the SL curve in Fig.~\ref{fig:toy_case}). When solving HJIs, the drawbacks of supervised learning are (1) its high data acquisition cost due to the need of solving BVPs and (2) its lack of generalization proof for solving PDEs. We hypothesize that these drawbacks can be addressed by combining supervised and self-supervised learning, since the latter only requires evaluating the HJI equations and is generalizable under continuity assumptions. 
% This makes Eq.~\eqref{eq:deepreach} a more computationally efficient way of learning values within each segment. 
% In the toy case, this leads to the HL curve which successfully approximates the ground truth.
% the latter only has proved convergence when player losses, the network, and the network gradient, are all continuous. Readers are referred to Sec.~\ref{sec:case} and Fig.~\ref{fig:complete info} for an example where self-supervised learning leads to poor generalization due to discontinuous player losses and raises safety concerns. 
This leads to the following hybrid method: Let $\mathcal{D}_s = \left \{\left(\textbf{x}_i^{(k)}, t^{(k)}, \nu_i^{(k)}, \nabla_{\textbf{x}_i}\nu_i^{(k)} \right) \text{ for } i=1,2 \right\}_{k=1}^K$ be a dataset derived from solving Eq.~\eqref{eq:pmp} with initial states sampled in $\mathcal{X}_1 \times \mathcal{X}_2$. The supervised loss is defined as:
% To this end, we propose a natural hybridization of the two approaches: We pre-train $(\hat{\nu}_1, \hat{\nu}_2)$ via Eq.~\eqref{eq:supervised}, and then refine the networks by minimizing $L_1 + L_2$. 
\cutequationup
\begin{equation}
\begin{aligned}
    L_2\left(\hat{\nu}_1, \hat{\nu}_2; \mathcal{D}_{\theta}\right) := & \sum_{k=1}^K \sum_{i=1}^2 \left| \hat{\nu}_i^{(k)} - \nu_i^{(k)} \right|\\
    & + C_2 \left \|\nabla_{\textbf{x}_i}\hat{\nu}_i^{(k)} - \nabla_{\textbf{x}_i} \nu_i^{(k)}\right \|,
    \label{eq:supervised}
\end{aligned}
\end{equation}
% \cutequationdown
where $C_2$ is a hyperparameter that balances the losses on value and its gradient. The hybrid method minimizes $L_1 + L_2$.

(2) \textit{Value hardening}: The second solution is to first introduce a surrogate differential equation, the continuous solution of which approximates the ground truth. Then we approximate the true solution by gradually ``hardening'' this surrogate. In the toy case, convergence of the solution can be achieved by gradually hardening a softened delta function (Fig.~\ref{fig:toy_case}). Nonetheless, a potential issue with this approach is that as we harden the delta function, the set of informative samples (that specifies non-zero $\nabla_x \nu$ in the toy case) reduces. Importance sampling becomes necessary to avoid convergence towards $\hat{\nu}=0$. In practice, it might improve the performance of value hardening by dynamically sampling states space, which leads to high PDE residuals~\cite{daw2022rethinking} 
% In practice, however, the importance of samples is unknown a priori.    

% \cutsectionup
\section{Case Studies}
% \cutsectiondown
\label{sec:case}
We compare generalization performance from self-supervised, supervised, hybrid, and value-hardening methods on two vehicle interaction simulation studies. The first considers an interaction between two players at an uncontrolled intersection, which leads to HJIs with coupled value functions defined on a 5D state space.
With proper coordinate transformation~\cite{chen2020high}, this case represents a broad range of challenging two-vehicle interactions including roundabouts and unprotected left turns. For completeness, 
%we study both a complete-information setting where players know each other's type, and an incomplete-information setting where player types are private and inferred during the interaction. 
we study both a complete- and an incomplete-information setting.
We then consider a  collision-avoidance case with higher state dimensionality similar to \cite{deepreach}. The difference is that here we encounter value discontinuity when players are required to both minimize effort and avoid collision. 

% We will use this case study to answer the following questions:
% \vspace{-0.05in}
% \begin{itemize}
%     \item Does the proposed algorithm achieve better generalization and safety performance than supervised and self-supervised approaches under the same computational budget?\cut
    
%     \item Are the generalization and safety performance sensitive to the choice of the network activation functions and the definition of training losses ($\phi({\cdot})$)?
%     \cut
%     %and the weighting of $L_1$ and $L_2$?  (not sure about this. since we say we pick Cs in order to make losses of equal magnitudes.)
    
%     \item Is it practical to decouple state and belief dynamics for the studied interactions?
% \end{itemize}
  
\subsection{Case 1: Uncontrolled intersection}  
\label{sec:uncontrolled intersection}
% \cutsubsectiondown

\begin{figure}
\centering
% \vspace{0.02in}
\includegraphics[width=0.95\linewidth]{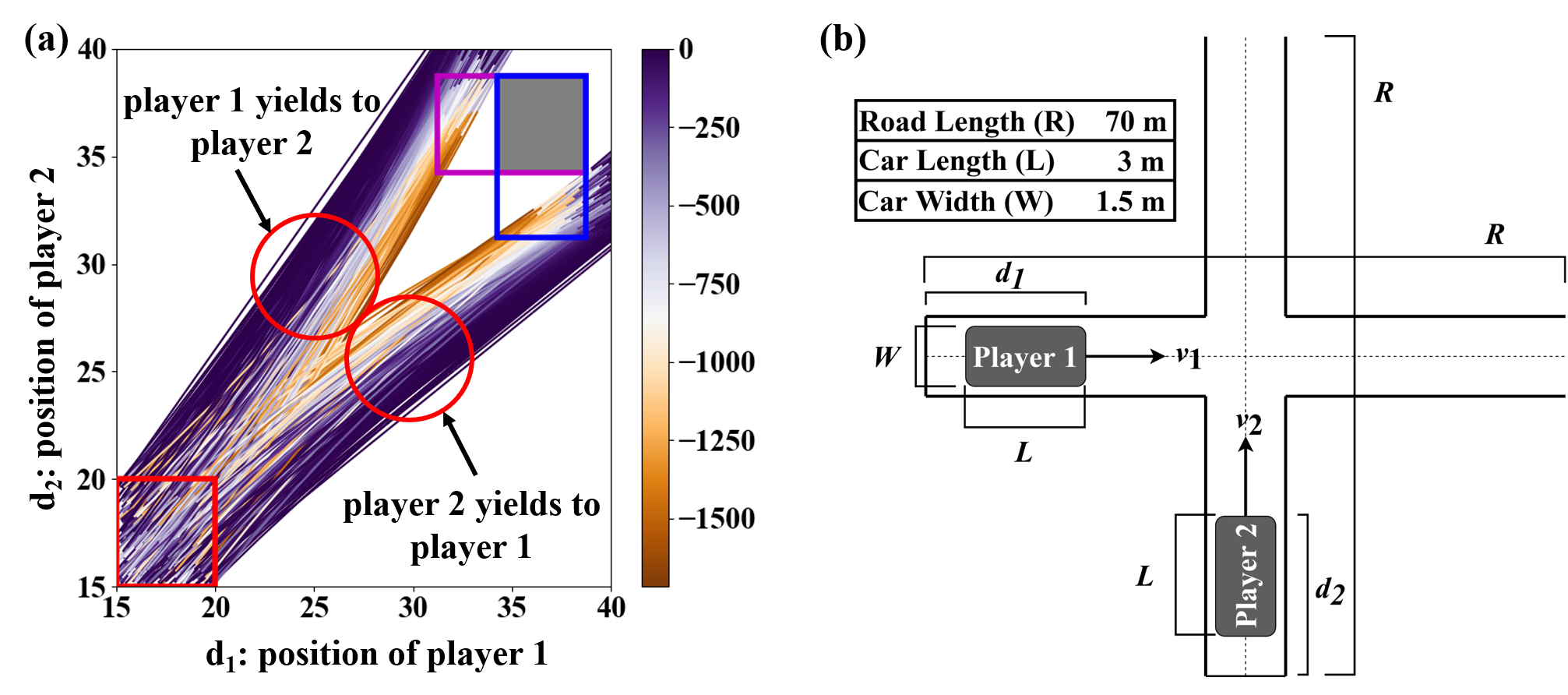}
\caption{(a) State trajectories of players projected to $(d_1, d_2)$. Solid
gray box: collision area from the perspective of aggressive players; hollow boxes (magenta for Player 1 and blue for Player 2): collision areas from the perspectives of non-aggressive players. Red box: sampling domain for initial states. Color: Actual values of Player 1. (b) Uncontrolled intersection setup.}
\label{fig:intersection case}
\vspace{-20pt}
\end{figure}

\textbf{Experiment setup.} 
% \textbf{Simulation setup.} 
The schematics of the interaction is shown in Fig.~\ref{fig:intersection case}, where Player $i$'s states are composed of two scalars representing its location ($d_i$) and speed ($v_i$): $x_i := (d_i, v_i)$. The shared dynamics follows $\dot{d}_i = v_i$ and $\dot{v}_i = u_i$, where $u_i \in [-5, 10] m/s^2$ is a scalar control input that represents the acceleration. The instantaneous loss considers the control effort and a collision penalty:
\begin{equation}
\begin{aligned}
    & l_i(\textbf{x}_i,u_i;\theta_i) = u_i^2 + b \sigma(d_i,\theta_i)\sigma(d_{-i},1),
    % ~\text{where} \\
    % & \sigma(d,\theta) = \left(1+\exp(-\gamma (d-R/2+\theta W/2))\right)^{-1}\left(1+\exp(\gamma (d-R/2-W/2-L))\right)^{-1}.
\end{aligned}
\end{equation}
where $\sigma(d,\theta) = 1$ iff $d \in [R/2 - \theta W/2, (R+W)/2 + L]$ or otherwise $\sigma(d,\theta) = 0$; $b=10^4$ sets a high loss for collision; 
% $\gamma=5$ is a shape parameter; 
$R$, $L$, and $W$ are the road length, %width, 
car length, and car width, respectively (see Fig.~\ref{fig:intersection case}). $\Theta = \{1, 5\}$ represents the aggressive (\texttt{a}) and non-aggressive (\texttt{na}) types of players, respectively. Note that the collision penalty is the source of discontinuity (or numerically, large Lipschitz constant, which contributes to the rapid value change in Fig.~\ref{fig:intersection case}).
% and softening is employed to stablize the BVP solver for solving Eq.~\eqref{eq:pmp}. 
The terminal loss is defined to incentivize players to move across the intersection and restore nominal speed: \begin{equation}
c_i(x) = -\alpha d_i(T) + (v_{i}(T)-\bar{v})^2,
\end{equation}
where $\alpha = 10^{-6}$, $\bar{v} = 18 m/s$, and $T = 3s$.

% \vspace{0.05in}
\begin{figure*}[!ht]
\centering
\includegraphics[width=\linewidth]{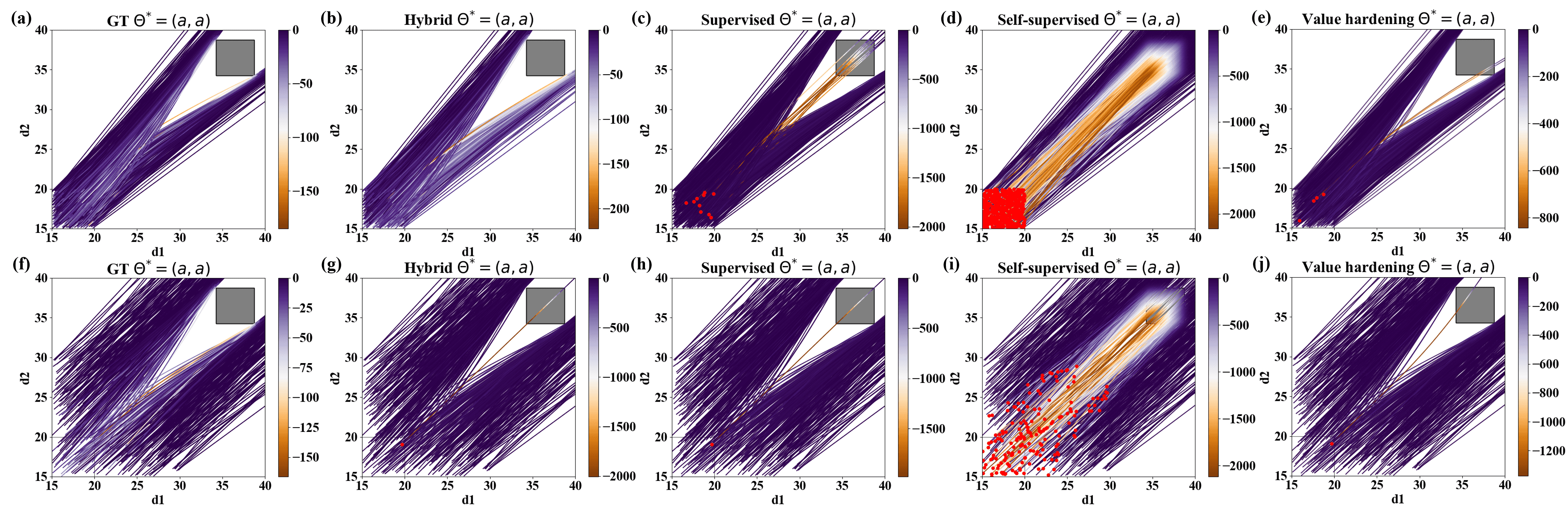}
\caption{(a), (f): Ground truth trajectories (projected to $d_1$-$d_2$) for $\mathcal{X}_{GT}$ and $\mathcal{X}_{XP}$, respectively. (b-e), (g-j): Trajectories generated using hybrid, supervised, self-supervised, and value-hardening methods under $\mathcal{X}_{GT}$ and $\mathcal{X}_{XP}$, respectively. Color: Actual equilibrial values of Player 1 along the trajectories. Trajectories with inevitable collisions are removed for clearer comparison on safety performance. Red dots represent initial states with \textit{avoidable} collisions.}
\vspace{-22pt}
\label{fig:complete info}
\end{figure*}

\textbf{Data.} For a fair comparison, data sizes for each learning method are chosen so that the total wall-clock computational costs for data acquisition and learning combined are similar across different methods. Note that we do not use computational complexity to measure the computational budget due to the different nature of computations involved in acquiring supervised data (which requires iterative BVP solving) and self-supervised data (which only requires random sampling). Specifically, all training sessions are around 300 minutes on one GeForce GTX 1080 Ti GPU with 11 GB memory\footnote{An exception is self-supervised learning, which is allowed to take around 500 minutes due to its poor generalization performance.}. Detailed data acquisition methods and data sizes are as follows:
For supervised learning, we generate 1.7k ground truth trajectories from initial states uniformly sampled in $\mathcal{X}_{GT} := [15, 20]m \times [18, 25]m/s$ and by solving Eq.~\eqref{eq:pmp}. Each trajectory contains 31 $\times$ 2 data-points (sampled with a time interval of $0.1s$ and for two players), which leads to a total of 105.4k data points. 
For self-supervised learning and its value-hardening variant, we sample 122k states uniformly in $\mathcal{X}_{HJ} := [15, 105]m \times [15, 32]m/s$. 
For hybrid learning, we sample 1k ground truth trajectories (62k data points) uniformly in $\mathcal{X}_{GT}$ and sample 60k states uniformly in $\mathcal{X}_{HJ}$.
We repeat the same sampling procedure for all four player type configurations (\texttt{a}, \texttt{a}), (\texttt{na}, \texttt{a}), (\texttt{a}, \texttt{na}), and (\texttt{na}, \texttt{na}). 
% The amount of data and states sampled for the three algorithms is determined so as to keep the total data procurement and training cost (in terms of wall-clock time) as close as possible for a fair comparison. 
% \textbf{Computational Cost.} Solving BVP is a computationally intense task, which is necessary to perform supervised learning. Both SL and HL methods require ground truth data. In the case of SL, we generate 1700 ground truth data, whereas in the case of HL, we generate 1000 ground truth data. 
% Table~\ref{tab:cost} shows the BVP solving time for these two cases. Furthermore, in order to ensure that the cost of training is same to train these models, we select the number of data points such that the training time is 300 minutes for each learning method. 
% Based on the results discussed above, HL is able to outperform others with much lower computational cost.

The choices of the state spaces to sample from are based on the following rationale: The ground truth trajectories are derived by sampling initial states from identical domains of the two players. This is because collision and near-collision cases, which are informative for the learning, often happen when players start with similar distances and speeds. Similarly, the %distance 
location range for supervised data ($[15, 20]m$) is calibrated so as to increase the chance of sampling informative trajectories within the specified time window. The speed range ($[18, 25]m/s$) is set based on nominal vehicle speed limits. For self-supervised learning, we use a space ($\mathcal{X}_{HJ}$) that approximately covers all states players can reach in $[0,T]$. Adaptive sampling is yet to be studied. 

%% Network Architecture
\textbf{Network Architecture.} Due to space limit, we only report results using fully-connected networks with 3 hidden layers, each containing 64 neurons, and with \texttt{tanh}, \texttt{relu}, or \texttt{sin} activations. Deeper or wider networks do not change our conclusions. \texttt{gelu} achieves similar performance as \texttt{tanh}. We normalize input data to $[-1, 1]$ to improve training convergence. Case 2 uses the same architecture.

\textbf{Training.} All training problems are solved using the Adam optimizer with a fixed learning rate of $2 \times 10^{-5}$.
For supervised learning, we train networks for a fixed 100k iterations. 
For self-supervised learning, we follow the curriculum learning method in \cite{deepreach} by first training the networks for 10k iterations using 1k uniformly sampled boundary states (at terminal time). The networks are then refined for 260k gradient descent steps, with states sampled from an expanding time window starting from the terminal. 
The value-hardening variant follows the same learning procedure but softens the collision penalty using sigmoid functions and gradually increases the shape parameter of the sigmoid to harden the penalty. For fair comparison, we use 5.4k training iterations for each hardening step for a total of 50 steps.
For the hybrid method, we pre-train the networks for 100k iterations using the supervised data, and combine the supervised data with states sampled from an expanding time window starting from the terminal time to minimize $L_1 + L_2$ for 100k iterations.

\begin{table}[!ht]
\vspace{0.05in}
\scriptsize
% \tiny
\centering
\caption{Generalization and safety performance on complete-information games. SL, SSL, HL, VH are for supervised, self-supervised, hybrid, value hardening methods, respectively.}
\label{table:complete info}
\begin{tabular}{lccccc}
\toprule
\multirow{2}{*}{Test} &
\multirow{2}{*}{Player} &
\multirow{2}{*}{Learning} &
\multicolumn{3}{c}{Metrics} \\ 
\cmidrule(lr){4-6}
Domain & Types & Method & $|\nu -\hat\nu|$ $\downarrow$	
  & $|u - \hat u|$ $\downarrow$ & Col.\% $\downarrow$\\ \midrule
  &  & SL & 0.57 &   0.12 $\pm$ 0.36 & 1.67\%  \\
  & (\texttt{a}, \texttt{a}) &  SSL & 3.39  & 0.96 $\pm$ 4.19 & 84.8\% \\
  &  & HL &  \textbf{0.46} & \textbf{0.09 $\pm$ 0.10} & \textbf{0.00\%} \\
  &  & VH &  4.17 & 0.34 $\pm$ 0.19 & 0.67\% \\
  \cmidrule(lr){3-6}
  &  & SL & 10.58 & 0.54 $\pm$ 3.92 & 4.50\%  \\
  $\mathcal{X}_{GT}$ & (\texttt{a}, \texttt{na}) &  SSL & 15.33 & 1.27 $\pm$ 7.16 & 83.3\%  \\ 
  &  & HL & \textbf{9.43} &  \textbf{0.49 $\pm$ 3.55}  & 3.50\% \\
  &  & VH &  79.35 & 1.10 $\pm$ 5.42 & \textbf{0.50\%} \\
  \cmidrule(lr){3-6}
  &  & SL & 3.49  & 0.10 $\pm$ 0.46 & 4.33\%  \\ 
  & (\texttt{na}, \texttt{na}) &  SSL & 114.67  & 1.88 $\pm$ 13.72 & 83.5\% \\
  &  & HL & \textbf{1.00} & \textbf{0.04 $\pm$ 0.03}  & \textbf{1.33\%} \\ 
  &  & VH &  21.76 & 0.34 $\pm$ 1.33 & 8.50\% \\
\midrule
  &  &  SL & 0.69 & 0.17 $\pm$ 0.28 & \textbf{0.20\%}  \\ 
  & (\texttt{a}, \texttt{a}) &  SSL & 1.54 & 0.37 $\pm$ 1.88  & 35.2\%  \\ 
  &  & HL & \textbf{0.41}  & \textbf{0.09 $\pm$ 0.08}  & \textbf{0.20\%} \\
  &  & VH &  2.03 & 0.20 $\pm$ 0.07 & \textbf{0.20\%} \\
  \cmidrule(lr){3-6}
  &  & SL & 19.01 & 0.56 $\pm$ 3.09 & 0.60\%  \\ 
  $\mathcal{X}_{XP}$ & (\texttt{a}, \texttt{na}) &  SSL & 19.57 & 0.58 $\pm$ 3.89 & 31.3\%  \\ 
  &  & HL & \textbf{17.39}  & \textbf{0.46 $\pm$ 3.17} & \textbf{0.10\%} \\
  &  & VH &  32.64 & 0.57 $\pm$ 2.71 & 0.20\% \\
  \cmidrule(lr){3-6}
  &  & SL & 4.25 & 0.30 $\pm$ 0.72 & 2.20\% \\
  & (\texttt{na}, \texttt{na}) &  SSL & 60.39 & 0.95 $\pm$ 7.31 & 36.0\%  \\ 
  &  & HL & \textbf{1.80} & \textbf{0.10 $\pm$ 0.12} & \textbf{0.00\%} \\ 
  &  & VH &  11.54 & 0.24 $\pm$ 0.68 & 6.40\% \\
\bottomrule
\end{tabular}
\vspace{-15pt}
\end{table}

\textbf{Results for complete-information games.} We generate a separate set of 600 ground truth trajectories with initial states in $\mathcal{X}_{GT}$ for each of the four player type configurations. For generalization performance, we measure the %relative mean square errors 
MAEs of value and control input predictions (denoted by $|\nu - \hat\nu|$ and $|u - \hat{u}|$ respectively) across the test trajectories.
% (variance of action predictions are given because we use action to compute closed-loop trajectories); 
For safety performance, we use the learned value networks to compute equilibrial Hamiltonian, and maximize the Hamiltonian to derive players' closed-loop control inputs. We report the percentage of \textit{avoidable} collisions. Initial states follow the test data. %Control inputs are computed by first auto-diffing the values and then maximizing the resultant Hamiltonian. 
Performance results are summarized in Table~\ref{table:complete info} and sample trajectories for the (\texttt{a}, \texttt{a}) case shown in Fig.~\ref{fig:complete info}. Performance of (\texttt{a}, \texttt{na}) and (\texttt{na}, \texttt{a}) are averaged due to the symmetry of players in these settings.

To examine the out-of-distribution performance of the models, we repeat the experiments in an expanded state space $\mathcal{X}_{XP}= [15, 30]m \times [18, 25]m/s$ and use 500 uniformly sampled initial states for tests. Results are summarized in the same table and figure. For both tests, the results show that the hybrid method achieves the best generalization and safety performance. Poor generalization of the self-supervised model is notable.

\begin{figure}[!ht]
\centering
\includegraphics[width=0.8\linewidth]{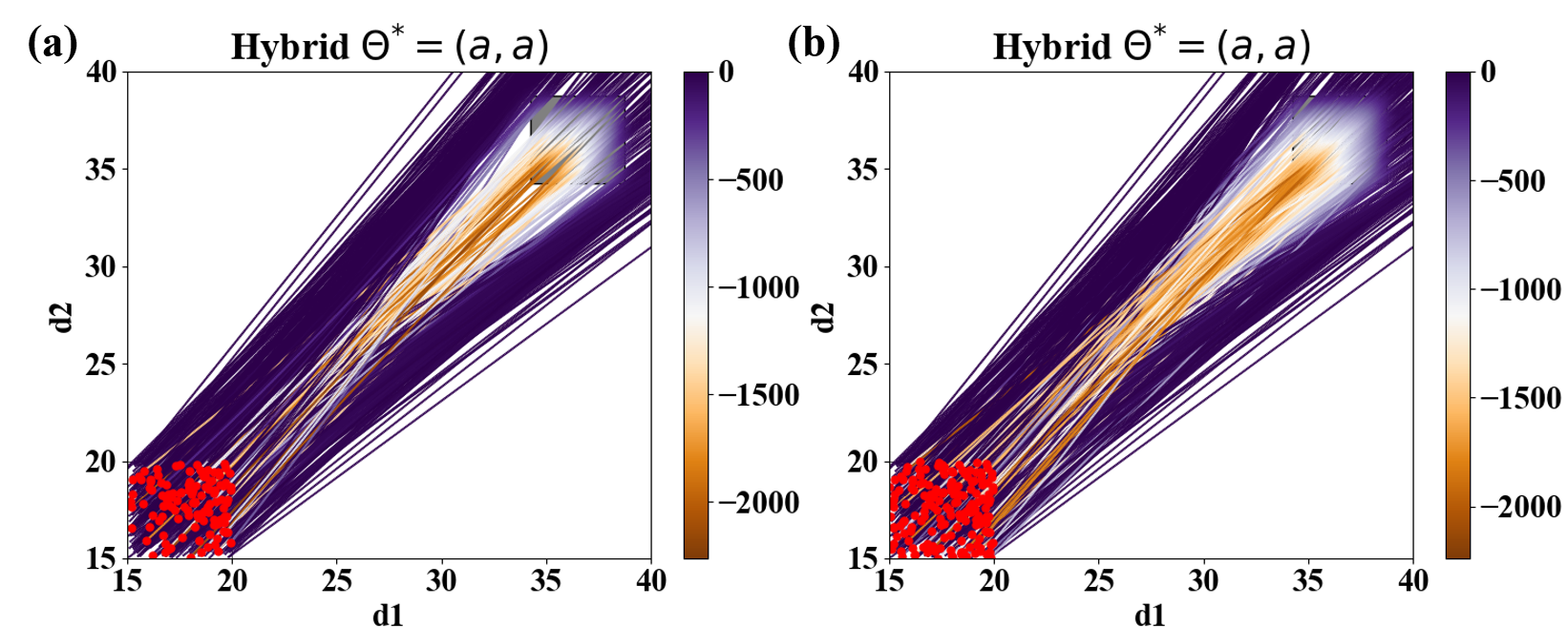}\\
\includegraphics[width=0.8\linewidth]{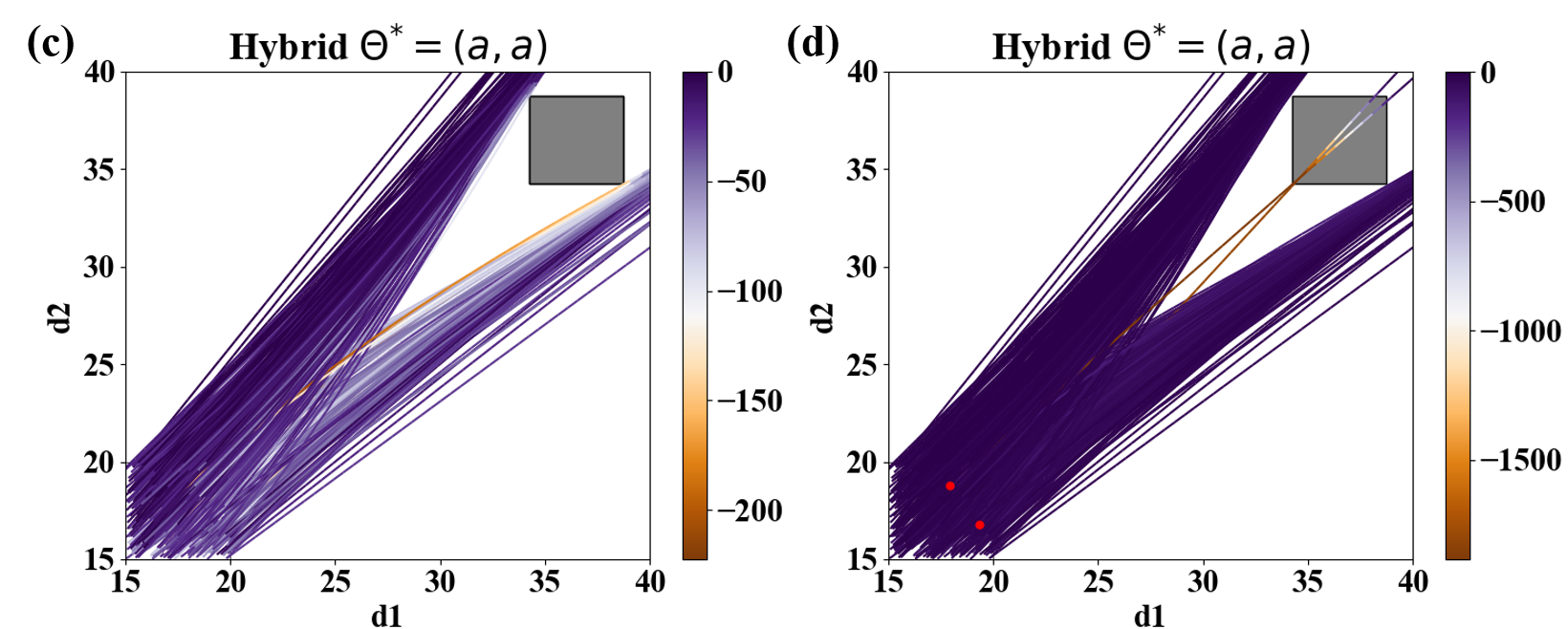}
\caption{Trajectories generated using neural networks with (a) \texttt{relu} and (b) \texttt{sin} activation functions and using $L^1$ for boundary norm (for \texttt{tanh}, refer to Fig.~\ref{fig:complete info}); trajectories generated using (c) $L^1$- and (d) $L^{2}$-norms for the boundary values and using $\texttt{tanh}$ for activation. All trajectories are based on hybrid learning.}
\vspace{-15pt}
\label{fig:ablation}
\end{figure}

\begin{table}[!ht]
\vspace{-5pt}
\centering
    \caption{Safety performance w/ different activation functions (w/ $L^1$) and boundary norms (w/ \texttt{tanh})}
    \label{table:activation}
\begin{tabular}{lccccc}
\toprule
\multirow{2}{*}{Method} & \multicolumn{3}{c}{Activation } & \multicolumn{2}{c}{Boundary Norm} \\ \cmidrule(lr){2-4}\cmidrule(lr){5-6} 
& \texttt{tanh}   & \texttt{relu}  & \texttt{sin}  & $L^1$   & $L^2$  \\ \midrule
Supervised  & 1.67\%  & 2.50\% & 19.5\% & - & - \\
Self-supervised  & 84.8\% & 84.0\%  & 84.7\% & -  & -   \\
Hybrid  & \textbf{0.0\%} & 19.8\%   & 28.7\%   & \textbf{0.0\%}  & 0.4\%  \\ \bottomrule
\end{tabular}
\vspace{-10pt}
\end{table}

\textbf{Ablation studies.} We perform ablation studies to understand the influences of activation functions and the norm of boundary loss on model performance. 
% Inspired by \citep{shin2020convergence, petersen2018optimal, siren}, we consider three different activation functions -- \texttt{tanh}, \texttt{relu}, and \texttt{sin} (in conjunction with the SIREN architecture). In the case of boundary value norm, we consider $L^1,$ and $L^2$ norms. 
Safety results are summarized in Table~\ref{table:activation} for player types (\texttt{a}, \texttt{a}) and using the hybrid learning method, with training and test in $\mathcal{X}_{GT}$. Corresponding trajectories are visualized in Fig.~\ref{fig:ablation}. Results show that (1) the choice of the activation function has a significant influence on the resultant models, with \texttt{tanh} outperforming \texttt{relu} and \texttt{sin}, and that (2) the choice of the boundary norm does not have significant influence. \textbf{Remarks}: (1) \texttt{relu} networks have proven convergence to piecewise smooth functions in a supervised setting~\cite{petersen2018optimal}. However, convergence in the self-supervised setting requires continuity of the network and its gradient~\cite{shin2020convergence}, which \texttt{relu} does not offer. This result is consistent with \cite{jagtap2020adaptive} where \texttt{relu} underperforms in solving PDEs. We conjecture that this contributes to the deteriorated performance of \texttt{relu} under the hybrid setting. We note, however, that smooth variants of \texttt{relu} such as \texttt{gelu} can achieve performance comparable to that of \texttt{tanh}. (2) \texttt{sin} does not perform well for Case 1, yet achieves comparable performance to \texttt{tanh} in Case 2. This result suggests that fine-tuning of the frequency parameter of \texttt{sin} is necessary and case-dependent.

% Please add the following required packages to your document preamble:
% \usepackage{multirow}
% \begin{table}[]
% \vspace{-5pt}
% \centering
%     \caption{Safety performance w/ different activation functions (w/ $L^1$) and boundary norms (w/ \texttt{tanh})}
%     \label{table:activation}
% \begin{tabular}{lccccc}
% \toprule
% \multirow{2}{*}{Method} & \multicolumn{3}{c}{Activation } & \multicolumn{2}{c}{Boundary Norm} \\ \cmidrule(lr){2-4}\cmidrule(lr){5-6} 
% & \texttt{tanh}   & \texttt{relu}  & \texttt{sin}  & $L^1$   & $L^2$  \\ \midrule
% Supervised  & 1.67\%  & 2.50\% & 19.5\% & - & - \\
% Self-supervised  & 84.8\% & 84.0\%  & 84.7\% & -  & -   \\
% Hybrid  & \textbf{0.0\%} & 19.8\%   & 28.7\%   & \textbf{0.0\%}  & 0.4\%  \\ \bottomrule
% \end{tabular}
% \vspace{-25pt}
% \end{table}

\textbf{Results for incomplete-information games.} We examine the advantage of the hybrid method when adopting value networks as closed-loop controllers for interactions in an incomplete-information setting.
% A driving parameter as a tuple $s = <\textbf{x}_0,\textbf{p}_0(\boldsymbol{\beta}),\boldsymbol{\theta}^{*},\textbf{l}>$ specifying the initial state, prior belief, true parameters, and estimation types and sample 150 different initial states from $\mathcal{X}= [15, 20]m \times [18, 25]m/s$. We use $a$, $na$, $n$, $ln$ for aggressive, non-aggressive, noisy, and less-noisy, respectively. Then we have a 4-by-4 matrix containing joint probabilities for the prior belief set $\mathcal{P}_0$. Each dimension of the matrix follows the order $(na,n), (na,ln), (a,n), (a,ln)$, e.g., the 1st row and 2nd column of the matrix represents $\Pr\left(\beta_1 = (na,n), \beta_2 = (na,ln)\right)$. 
To constrain the scope of tests, we 
% assume that agents are mostly rational ($\Pr(\lambda_{1,2} = ln) = 0.8$), and 
explore two settings of initial beliefs: Everyone believes that everyone is (1) most likely non-aggressive ($p_i := \Pr(\theta_i = a) = 0.2$), or (2) most likely aggressive ($p_i = 0.8$ for $i=1,2$). The common belief assumption indicates that Player $i$ knows about Player $j$'s belief about $i$. Also note that players' initial beliefs can mismatch with their true types.
% This reduces $\mathcal{P}_0$ to $\{p_0^{na}, p_0^{a}\}$, where $p_0^{na}$ and $p_0^{a}$ are the common priors where everyone is believed to be non-aggressive and aggressive, respectively. We set $\mathcal{L}=\{(e,e),(ne,ne)\}$ where $e$ stands for ``empathetic'' and $ne$ for ``non-empathetic
% In the simulation, we set the time interval as $0.05s$, so the interaction trajectories can be computed for each driving scenario $s$ through HL and SL. 
We model players to continuously perform Bayes updates on their beliefs based on observations, and determine the next control inputs based on the value networks that correspond to the most likely player types according to their current beliefs (see \cite{chen2021shall} for details on belief update and control). This model allows us to simulate state and belief dynamics in turn, although the values are approximated based on complete-information games and are thus decoupled from the belief dynamics. 
Table~\ref{table:inference} compares safety performance of incomplete-information interactions using the same initial states as previous tests and values learned by the hybrid and the supervised methods. The former outperforms the latter across all belief settings.

\begin{table}[h!]
\vspace{-5pt}
\centering
    \caption{Safety performance in uncontrolled intersections with incomplete information}
    \label{table:inference}
    \begin{tabular}{lccc} 
      \toprule
      True types & Initial common belief $(p_1, p_2)$ & Hybrid & Supervised\\\midrule
      (\texttt{a},\texttt{a}) & (0.8,0.8) & $\textbf{0.00\%}$ & $\textbf{0.00\%}$\\
      (\texttt{a},\texttt{a}) & (0.2,0.2) & $\textbf{2.00\%}$ & $8.00\%$\\
      (\texttt{na},\texttt{na}) & (0.8,0.8) & $\textbf{2.00\%}$ & $2.67\%$\\
      (\texttt{na},\texttt{na}) & (0.2,0.2) & $\textbf{0.67\%}$ & $6.67\%$\\
      \bottomrule
    \end{tabular}
    \vspace{-15pt}
\end{table}

\begin{figure*}[!ht]
\centering
\includegraphics[width=\linewidth]{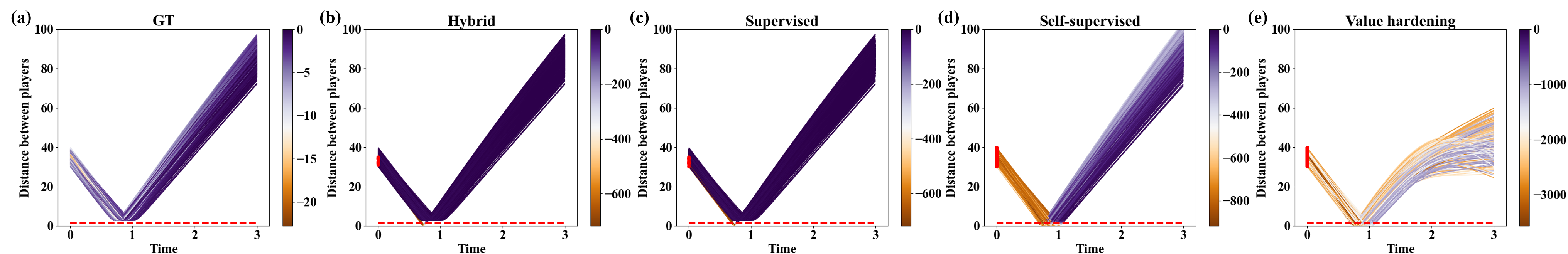}
\caption{(a): Ground truth distance between players over time for $\mathcal{X}_{GT}$. (b-e): Distance between players over time using hybrid, supervised, self-supervised, and value hardening models under $\mathcal{X}_{GT}$, respectively. %Color: Actual values of Player 1. Inevitable collisions are removed for clearer comparison on safety performance. Red dots represent initial states with \textit{avoidable} collisions. 
Red dashed line represents the threshold distance for collision.}
\label{fig:HD_case}
\vspace{-10pt}
\end{figure*}

\begin{figure*}[!ht]
\centering
\includegraphics[width=\linewidth]{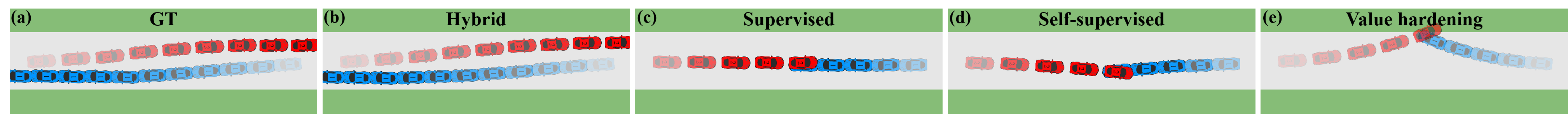}
\caption{Collision avoidance visualization: (a): Ground truth safe trajectory. Transparency reduces along time. (b-e): Trajectories generated using hybrid, supervised, self-supervised, and value hardening models, respectively.}
\label{fig:HD_visualization}
\vspace{-20pt}
\end{figure*}

\subsection{Case 2: Collision avoidance} 
\label{sec:narrow-road}
% \cutsubsectiondown

\begin{figure}[!ht]
\centering
\includegraphics[width=0.75\linewidth]{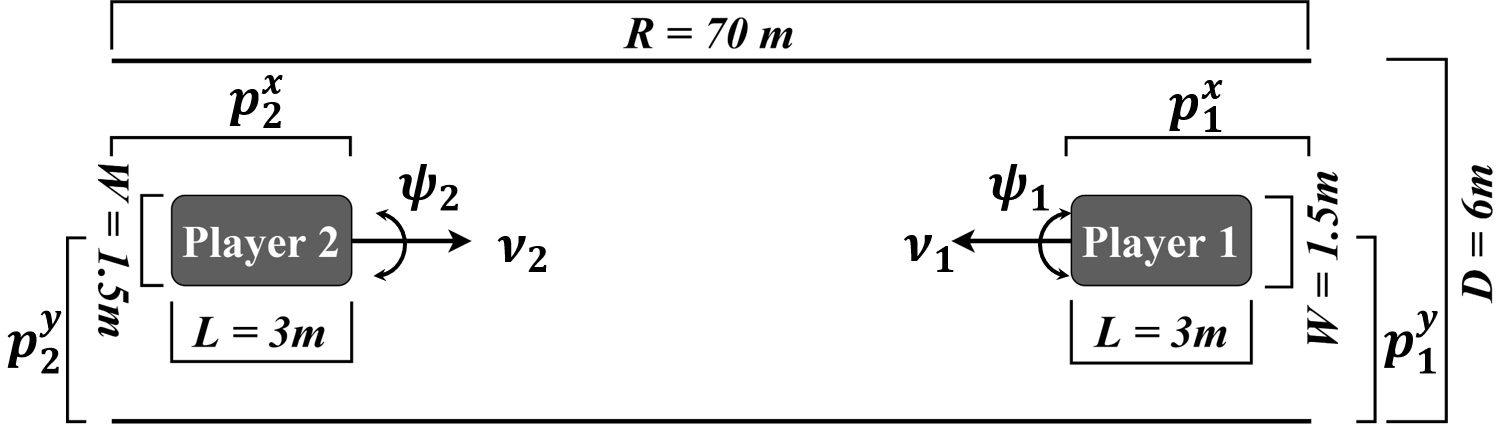}
\caption{Collision avoidance setup with two players.}
\label{fig:narrow road case}
\vspace{-10pt}
\end{figure}

\textbf{Experiment setup.} 
% \textbf{Simulation setup.} 
The schematics is shown in Fig.~\ref{fig:narrow road case}, where Player $i$'s states are composed of its location ($p^{x}_i$, $p^{y}_i$), orientation ($\psi_i$) and speed ($v_i$): $x_i := [p^{x}_i, p^{y}_i, \psi_i, v_i]^T$. We consider a unicycle model as the dynamics: 
\begin{eqnarray}
\begin{aligned}
\left[
\begin{array}{c}
    \dot p^{x}_i \\
    \dot p^{y}_i \\
    \dot \psi_i \\
    \dot v_i \\
\end{array}
\right]
=
\left[
\begin{array}{c}
    v_i\cos(\psi_i) \\
    v_i\sin(\psi_i) \\
    \omega_i \\
    u_i \\
\end{array}
\right]
\end{aligned}
\end{eqnarray} 
where $\omega_i \in [-1, 1] rad/s$ and $u_i \in [-5, 10] m/s^2$ are control inputs that represent angular velocity and acceleration, respectively. The instantaneous loss considers the control effort and the collision penalty:
\begin{equation}
\begin{aligned}
    & l_i(\textbf{x}_i, \omega_i, u_i) = k\omega_i^2 + u_i^2 + b\sigma(\textbf{x}_i,\eta),
\end{aligned}
\end{equation}
where the penalty is $\sigma(\textbf{x}_i,\eta) = \left(1+\exp(\gamma (\sqrt{((R - p^{x}_2) - p^{x}_1)^2 + (p^{y}_2 - p^{y}_1)^2} - \eta)\right)^{-1}$. $b=10^4$ sets a high loss for collision; $\gamma=5$ is a shape parameter; $R$ is road length; $k=100$ balances the control effort on turning and acceleration; $\eta=1.5 m$ is a threshold for collision. The terminal loss encourages players to move along the lane and restore nominal speed: 
\begin{equation}
c_i(x) = -\alpha p^x_{i}(T) + (v_{i}(T)-\bar{v})^2 + (p^y_{i}(T)-\bar{p}^{y})^2,
\end{equation}
where $\alpha = 10^{-6}$, $\bar{v} = 18 m/s$, $\bar{p}^{y} = 3 m$, and $T = 3s$.

% \begin{wrapfigure}{r}{0.3\textwidth}
% \centering
% \vspace{-10pt}
% \includegraphics[width=1.0\linewidth]{rebuttal/narrow_road.png}
% \caption{Narrow-road setup with two players.}
% \vspace{-10pt}
% \label{fig:narrow road case}
% \end{wrapfigure}

\textbf{Data.} For supervised learning, we generate 1.45k ground truth trajectories from initial states uniformly sampled in $\mathcal{X}_{GT} := [15, 20]m \times [2.25, 3.75]m \times [-\pi/180, \pi/180] rad \times [18, 25]m/s$,
% Each trajectory contains 31 $\times$ 2 data-points (sampled with a time interval of $0.1s$ and for two players), 
which leads to a total of 89.9k data points. 
For self-supervised learning and its value-hardening variant, we sample 122k states uniformly in $\mathcal{X}_{HJ} := [15, 90]m \times [0, 6]m \times [-0.15, 0.18]rad \times [18, 25]m/s$. 
For hybrid learning, we generate 1k ground truth trajectories (62k data points) in $[15, 20]m \times [2.25, 3.75]m \times [-\pi/180, \pi/180] rad \times [18, 25]m/s$ and sample 60k states uniformly in $[15, 90]m \times [0, 6]m \times [-0.15, 0.18]rad \times [18, 25]m/s$. 
% The rationale for the varying amount of data is the same as for Case 1.

% \begin{wrapfigure}{r}{0.3\textwidth}
% \centering
% \vspace{-10pt}
% \includegraphics[width=1.0\linewidth]{rebuttal/narrow_road.png}
% \caption{Narrow-road setup with two players.}
% \vspace{-10pt}
% \label{fig:narrow road case}
% \end{wrapfigure}

%% Network Architecture
% \textbf{Network Architecture.} Same as Case 1. 

% For all experiments, we use a fully-connected network with 3 hidden layers, each containing 64 neurons. We use \texttt{tanh} as the activation functions. Following \citep{deepreach}, we normalize input data to $[-1, 1]$ to improve training convergence.
\textbf{Training.} 
%Supervised learning uses 100k training iterations. 
Self-supervised learning adopts pretraining of networks for 10k iterations using 1k uniformly sampled boundary states and trains the network for 350k iterations. The value-hardening variant uses 7.2k training iterations for each hardening step with a total of 50 steps for fair comparison. The rest of the settings follows Case 1.

\textbf{Results.} We use a separate test set of 600 ground truth collision-free trajectories with initial states drawn from $\mathcal{X}_{GT}$. 
% For safety performance, we use the learned value networks as closed-loop controllers to generate complete-information narrow-road case study, and report the percentage of \textit{avoidable} collisions. 
% Initial states follow the test data. Control inputs are computed by auto-diffing the values and maximizing the resultant Hamiltonian. 
Performance results are summarized in Table~\ref{table:HD_safety}. Distance between players during the interactions are visualized in Fig.~\ref{fig:HD_case}. Similar to Case 1, hybrid learning outperforms other methods. Value-hardening fails to generalize in this higher-dimensional case. Comparing with Case 1, we conjecture that the failure is caused by a decreasing probability of capturing informative samples (where rapid value changes happen) as the state dimension increases.  Fig.~\ref{fig:HD_visualization} illustrates one initial state where the hybrid method induces a safe interaction while other baselines have collisions. 

\begin{table}[h!]
\centering
    \caption{Safety performance w/ activation function \texttt{tanh} under $\mathcal{X}_{GT}$}
    \label{table:HD_safety}
    \begin{tabular}{ccccc} 
      \toprule
      & \textbf{HL} & \textbf{SL} & \textbf{SSL} & \textbf{VH}\\\midrule
      Col.\% & \textbf{1.67}\% & 2.17\% & 98.8\% & 95.7\%\\
      \bottomrule
    \end{tabular}
    \vspace{-10pt}
\end{table}

% \begin{figure*}[!h]
% \centering
% \includegraphics[width=\linewidth]{rebuttal/HD_case.png}
% \caption{(a): Ground truth distance between players over time for $\mathcal{X}_{GT}$. (b-d): Distance between players over time using hybrid, supervised, and self-supervised models under $\mathcal{X}_{GT}$, respectively. Color: Actual values of Player 1. Inevitable collisions are removed for clearer comparison on safety performance. Red dots represent initial states with \textit{avoidable} collisions. Red dashed line represents the threshold distance for collision.}
% \label{fig:HD_case}
% \vspace{-10pt}
% \end{figure*}

% \begin{figure*}[!h]
% \centering
% \includegraphics[width=\linewidth]{rebuttal/HD_visualization.png}
% \caption{Collision avoidance visualization: (a): Ground truth safe trajectory. Transparency reduces along time. (b-d): Trajectories generated using hybrid, supervised, and self-supervised models, respectively.}
% \label{fig:HD_visualization}
% \vspace{-10pt}
% \end{figure*}

\section{Conclusions}
% \cutsectiondown
\label{sec:conclusion}
We proposed a hybrid learning method that combines the advantages of supervised and self-supervised learning for approximating discontinuous value functions as solutions to two-player general-sum differential games. Using two vehicles interaction cases, we showed that the proposed method leads to better generalization and safety performance than purely supervised and self-supervised learning when using the same computational cost. 
%This new algorithm can offset the limitation of SL and SSL, and outperform the latter two algorithms within the same computational budget. 
In essence, our algorithm allows the value network to first adapt to the structure of the value function through supervised data, and then uses the underlying governing equations to satisfy the PDE in the entire state space.
We also tested the idea of value hardening, and showed that without informative supervised data, this method does not guarantee convergence to the true values when discontinuity exists.
Lastly, we empirically showed the importance of continuous differentiability of activation functions in solving PDEs with deep learning. It is worth noting that state constraints in differential games, which are the cause of large penalties that motivated this study, can also be incorporated into a different HJ formulation with an extended state space through the epigraph technique~\cite{altarovici2013general,lee2022safety}. This leads to value approximation in a higher dimensional state space yet allows the resulting value to be continuous on that space. Our future work will investigate the effectiveness of self-supervised learning for this new formulation of HJ equations. 
\bibliography{icra2023}

% Generated by IEEEtran.bst, version: 1.14 (2015/08/26)
\begin{thebibliography}{10}
\providecommand{\url}[1]{#1}
\csname url@samestyle\endcsname
\providecommand{\newblock}{\relax}
\providecommand{\bibinfo}[2]{#2}
\providecommand{\BIBentrySTDinterwordspacing}{\spaceskip=0pt\relax}
\providecommand{\BIBentryALTinterwordstretchfactor}{4}
\providecommand{\BIBentryALTinterwordspacing}{\spaceskip=\fontdimen2\font plus
\BIBentryALTinterwordstretchfactor\fontdimen3\font minus
  \fontdimen4\font\relax}
\providecommand{\BIBforeignlanguage}[2]{{%
\expandafter\ifx\csname l@#1\endcsname\relax
\typeout{** WARNING: IEEEtran.bst: No hyphenation pattern has been}%
\typeout{** loaded for the language `#1'. Using the pattern for}%
\typeout{** the default language instead.}%
\else
\language=\csname l@#1\endcsname
\fi
#2}}
\providecommand{\BIBdecl}{\relax}
\BIBdecl

\bibitem{viscosity}
M.~G. Crandall and P.-L. Lions, ``Viscosity solutions of hamilton-jacobi
  equations,'' \emph{Transactions of the American mathematical society}, vol.
  277, no.~1, pp. 1--42, 1983.

\bibitem{mitchell2003}
I.~M. Mitchell and C.~J. Tomlin, ``Overapproximating reachable sets by
  hamilton-jacobi projections,'' \emph{journal of Scientific Computing},
  vol.~19, no.~1, pp. 323--346, 2003.

\bibitem{deepreach}
S.~Bansal and C.~J. Tomlin, ``{DeepReach}: A deep learning approach to
  high-dimensional reachability,'' in \emph{2021 IEEE International Conference
  on Robotics and Automation (ICRA)}.\hskip 1em plus 0.5em minus 0.4em\relax
  IEEE, 2021, pp. 1817--1824.

\bibitem{shin2020convergence}
Y.~Shin, J.~Darbon, and G.~E. Karniadakis, ``On the convergence of physics
  informed neural networks for linear second-order elliptic and parabolic type
  pdes,'' \emph{arXiv preprint arXiv:2004.01806}, 2020.

\bibitem{mitchell2005}
I.~M. Mitchell, A.~M. Bayen, and C.~J. Tomlin, ``A time-dependent
  hamilton-jacobi formulation of reachable sets for continuous dynamic games,''
  \emph{IEEE Transactions on automatic control}, vol.~50, no.~7, pp. 947--957,
  2005.

\bibitem{mangasarian1966sufficient}
O.~L. Mangasarian, ``Sufficient conditions for the optimal control of nonlinear
  systems,'' \emph{SIAM Journal on control}, vol.~4, no.~1, pp. 139--152, 1966.

\bibitem{jagtap2020adaptive}
A.~D. Jagtap, K.~Kawaguchi, and G.~E. Karniadakis, ``Adaptive activation
  functions accelerate convergence in deep and physics-informed neural
  networks,'' \emph{Journal of Computational Physics}, vol. 404, p. 109136,
  2020.

\bibitem{autodiff}
A.~G. Baydin, B.~A. Pearlmutter, A.~A. Radul, and J.~M. Siskind, ``Automatic
  differentiation in machine learning: a survey,'' \emph{Journal of Marchine
  Learning Research}, vol.~18, pp. 1--43, 2018.

\bibitem{bellman1965dynamic}
R.~Bellman and R.~E. Kalaba, \emph{Dynamic programming and modern control
  theory}.\hskip 1em plus 0.5em minus 0.4em\relax Citeseer, 1965, vol.~81.

\bibitem{bsde}
J.~Han, A.~Jentzen, and E.~Weinan, ``Solving high-dimensional partial
  differential equations using deep learning,'' \emph{Proceedings of the
  National Academy of Sciences}, vol. 115, no.~34, pp. 8505--8510, 2018.

\bibitem{han2020convergence}
J.~Han and J.~Long, ``Convergence of the deep bsde method for coupled fbsdes,''
  \emph{Probability, Uncertainty and Quantitative Risk}, vol.~5, no.~1, pp.
  1--33, 2020.

\bibitem{ito2021neural}
K.~Ito, C.~Reisinger, and Y.~Zhang, ``A neural network-based policy iteration
  algorithm with global h2-superlinear convergence for stochastic games on
  domains,'' \emph{Foundations of Computational Mathematics}, vol.~21, no.~2,
  pp. 331--374, 2021.

\bibitem{zinkevich2007regret}
M.~Zinkevich, M.~Johanson, M.~Bowling, and C.~Piccione, ``Regret minimization
  in games with incomplete information,'' \emph{Advances in neural information
  processing systems}, vol.~20, 2007.

\bibitem{perolat2022mastering}
J.~Perolat, B.~De~Vylder, D.~Hennes, E.~Tarassov, F.~Strub, V.~de~Boer,
  P.~Muller, J.~T. Connor, N.~Burch, T.~Anthony \emph{et~al.}, ``Mastering the
  game of stratego with model-free multiagent reinforcement learning,''
  \emph{Science}, vol. 378, no. 6623, pp. 990--996, 2022.

\bibitem{tammelin2015solving}
O.~Tammelin, N.~Burch, M.~Johanson, and M.~Bowling, ``Solving heads-up limit
  texas hold'em,'' in \emph{Twenty-fourth international joint conference on
  artificial intelligence}, 2015.

\bibitem{facebook_poker}
N.~Brown, A.~Bakhtin, A.~Lerer, and Q.~Gong, ``Combining deep reinforcement
  learning and search for imperfect-information games,'' \emph{Advances in
  Neural Information Processing Systems}, vol.~33, pp. 17\,057--17\,069, 2020.

\bibitem{foerster_learning_2017}
J.~N. Foerster, R.~Y. Chen, M.~Al-Shedivat, S.~Whiteson, P.~Abbeel, and
  I.~Mordatch, ``Learning with {Opponent}-{Learning} {Awareness},''
  \emph{arXiv:1709.04326 [cs]}, Sep. 2017, arXiv: 1709.04326.

\bibitem{sadigh_planning_2018}
D.~Sadigh, N.~Landolfi, S.~S. Sastry, S.~A. Seshia, and A.~D. Dragan,
  ``\BIBforeignlanguage{en}{Planning for cars that coordinate with people:
  leveraging effects on human actions for planning and active information
  gathering over human internal state},''
  \emph{\BIBforeignlanguage{en}{Autonomous Robots}}, vol.~42, no.~7, pp.
  1405--1426, Oct. 2018.

\bibitem{kwon2020humans}
M.~Kwon, E.~Biyik, A.~Talati, K.~Bhasin, D.~P. Losey, and D.~Sadigh, ``When
  humans aren't optimal: Robots that collaborate with risk-aware humans,'' in
  \emph{Proceedings of the 2020 ACM/IEEE International Conference on
  Human-Robot Interaction}, 2020, pp. 43--52.

\bibitem{schwarting2019social}
W.~Schwarting, A.~Pierson, J.~Alonso-Mora, S.~Karaman, and D.~Rus, ``Social
  behavior for autonomous vehicles,'' \emph{Proceedings of the National Academy
  of Sciences}, vol. 116, no.~50, pp. 24\,972--24\,978, 2019.

\bibitem{zahedi2022seeking}
Z.~Zahedi, A.~Khayatian, M.~M. Arefi, and S.~Yin, ``Seeking nash equilibrium in
  non-cooperative differential games,'' \emph{Journal of Vibration and
  Control}, p. 10775463221122120, 2022.

\bibitem{li2022off}
J.~Li, Z.~Xiao, J.~Fan, T.~Chai, and F.~L. Lewis, ``Off-policy q-learning:
  Solving nash equilibrium of multi-player games with network-induced delay and
  unmeasured state,'' \emph{Automatica}, vol. 136, p. 110076, 2022.

\bibitem{nikolaidis2017human}
S.~Nikolaidis, D.~Hsu, and S.~Srinivasa, ``Human-robot mutual adaptation in
  collaborative tasks: Models and experiments,'' \emph{The International
  Journal of Robotics Research}, vol.~36, no. 5-7, pp. 618--634, 2017.

\bibitem{sun2018probabilistic}
L.~Sun, W.~Zhan, and M.~Tomizuka, ``Probabilistic prediction of interactive
  driving behavior via hierarchical inverse reinforcement learning,'' in
  \emph{2018 21st International Conference on Intelligent Transportation
  Systems (ITSC)}.\hskip 1em plus 0.5em minus 0.4em\relax IEEE, 2018, pp.
  2111--2117.

\bibitem{peng2019bayesian}
C.~Peng and M.~Tomizuka, ``Bayesian persuasive driving,'' in \emph{2019
  American Control Conference (ACC)}.\hskip 1em plus 0.5em minus 0.4em\relax
  IEEE, 2019, pp. 723--729.

\bibitem{wang2019enabling}
Y.~Wang, Y.~Ren, S.~Elliott, and W.~Zhang, ``Enabling courteous vehicle
  interactions through game-based and dynamics-aware intent inference,''
  \emph{IEEE Transactions on Intelligent Vehicles}, vol.~5, no.~2, pp.
  217--228, 2019.

\bibitem{fridovich2020confidence}
D.~Fridovich-Keil, A.~Bajcsy, J.~F. Fisac, S.~L. Herbert, S.~Wang, A.~D.
  Dragan, and C.~J. Tomlin, ``Confidence-aware motion prediction for real-time
  collision avoidance1,'' \emph{The International Journal of Robotics
  Research}, vol.~39, no. 2-3, pp. 250--265, 2020.

\bibitem{chen2021shall}
Y.~Chen, L.~Zhang, T.~Merry, S.~Amatya, W.~Zhang, and Y.~Ren, ``When shall i be
  empathetic? the utility of empathetic parameter estimation in multi-agent
  interactions,'' in \emph{2021 IEEE International Conference on Robotics and
  Automation (ICRA)}.\hskip 1em plus 0.5em minus 0.4em\relax IEEE, 2021, pp.
  2761--2767.

\bibitem{fridovich2020efficient}
D.~Fridovich-Keil, E.~Ratner, L.~Peters, A.~D. Dragan, and C.~J. Tomlin,
  ``Efficient iterative linear-quadratic approximations for nonlinear
  multi-player general-sum differential games,'' in \emph{2020 IEEE
  international conference on robotics and automation (ICRA)}.\hskip 1em plus
  0.5em minus 0.4em\relax IEEE, 2020, pp. 1475--1481.

\bibitem{starr1969nonzero}
A.~W. Starr and Y.-C. Ho, ``Nonzero-sum differential games,'' \emph{Journal of
  optimization theory and applications}, vol.~3, no.~3, pp. 184--206, 1969.

\bibitem{singarcs}
E.~Cristiani and P.~Martinon, ``Initialization of the shooting method via the
  hamilton-jacobi-bellman approach,'' \emph{Journal of Optimization Theory and
  Applications}, vol. 146, no.~2, pp. 321--346, 2010.

\bibitem{daw2022rethinking}
A.~Daw, J.~Bu, S.~Wang, P.~Perdikaris, and A.~Karpatne, ``Rethinking the
  importance of sampling in physics-informed neural networks,'' \emph{arXiv
  preprint arXiv:2207.02338}, 2022.

\bibitem{chen2020high}
X.~Chen, Z.~Li, Y.~Yang, L.~Qi, and R.~Ke, ``High-resolution vehicle trajectory
  extraction and denoising from aerial videos,'' \emph{IEEE Transactions on
  Intelligent Transportation Systems}, vol.~22, no.~5, pp. 3190--3202, 2020.

\bibitem{petersen2018optimal}
P.~Petersen and F.~Voigtlaender, ``Optimal approximation of piecewise smooth
  functions using deep relu neural networks,'' \emph{Neural Networks}, vol.
  108, pp. 296--330, 2018.

\bibitem{altarovici2013general}
A.~Altarovici, O.~Bokanowski, and H.~Zidani, ``A general hamilton-jacobi
  framework for non-linear state-constrained control problems,'' \emph{ESAIM:
  Control, Optimisation and Calculus of Variations}, vol.~19, no.~2, pp.
  337--357, 2013.

\bibitem{lee2022safety}
D.~Lee, ``Safety-guaranteed autonomy under uncertainty,'' Ph.D. dissertation,
  UC Berkeley, 2022.

\end{thebibliography}
\bibliographystyle{IEEEtran}
\end{document}